\documentclass{article}

\usepackage{PRIMEarxiv}

\usepackage[utf8]{inputenc} 
\usepackage[T1]{fontenc}    
\usepackage{hyperref}       
\usepackage{url}            
\usepackage{booktabs}       
\usepackage{amsmath,amsfonts,amssymb}       
\usepackage{nicefrac}       
\usepackage{microtype}      
\usepackage{lipsum}
\usepackage{fancyhdr}       
\usepackage{graphicx}       
\graphicspath{{media/}}     

\usepackage{algorithmic}
\usepackage{array}
\usepackage[caption=false,font=normalsize,labelfont=sf,textfont=sf]{subfig}
\usepackage{textcomp}
\usepackage{stfloats}
\usepackage{verbatim}
\usepackage{cite}
\usepackage{bm}
\usepackage{multirow}
\usepackage{adjustbox}
\usepackage{hyperref}
\usepackage[ruled]{algorithm2e}
\usepackage{xcolor}
\hypersetup{
    colorlinks
}

\title{\emph{FedER}: Federated Learning through Experience Replay and Privacy-Preserving Data Synthesis
}

\author{
  M. Pennisi$^*$, F. Proietto Salanitri$^*$, G. Bellitto, S. Palazzo, C. Spampinato\\
  Department of Electrical, Electronics and Computer Engineering\\
  University of Catania \\
  Catania, Italy\\
  \texttt{\{matteo.pennisi, federica.proiettosalanitri, giovanni.bellitto\}@phd.unict.it} \\
  \texttt{\{simone.palazzo, concetto.spampinato\}@unict.it} \\
   \And
  B. Casella, M. Aldinucci\\
  Department of Computer Science \\
  University of Turin \\
  Turin, Italy\\
  \texttt{\{bruno.casella, marco.aldinucci\}@unito.it} \\
}

\newcommand{\mytilde}{\raise.17ex\hbox{$\scriptstyle\mathtt{\sim}$}}
\DeclareMathOperator*{\argmin}{arg\,min}

\newcommand{\red}[1]{\textcolor{black}{#1}}

\DeclareMathOperator{\EX}{\mathbb{E}}

\def\ww{\mathbf{w}}
\def\xx{\mathbf{x}}
\def\yy{\mathbf{y}}
\def\zz{\mathbf{z}}

\def\bB{\mathcal{B}}

\def\dD{\mathcal{D}}

\def\gG{\mathcal{G}}

\def\lL{\mathcal{L}}
\def\mM{\mathcal{M}}

\def\tT{\mathcal{T}}

\def\xX{\mathcal{X}}
\def\yY{\mathcal{Y}}
\def\zZ{\mathcal{Z}}

\def\Ee{\mathbb{E}}

\begin{document}
\maketitle
\def\thefootnote{*}\footnotetext{Equal contribution}\def\thefootnote{\arabic{footnote}}

\begin{abstract}
In the medical field, multi-center collaborations are often sought to yield more generalizable findings by leveraging the heterogeneity of patient and clinical data. However, recent privacy regulations hinder the possibility to share data, and consequently, to come up with machine learning-based solutions that support diagnosis and prognosis.
Federated learning (FL) aims at sidestepping this limitation by bringing AI-based solutions to data owners and only sharing local AI models, or parts thereof, that need then to be aggregated. 
However, most of the existing federated learning solutions are still at their infancy and show several shortcomings, from the lack of a reliable and effective aggregation scheme able to retain the knowledge learned locally to weak privacy preservation as real data may be reconstructed from model updates.
Furthermore, the majority of these approaches, especially those dealing with medical data, relies on a centralized distributed learning strategy that poses robustness, scalability and trust issues.
In this paper we present a federated and decentralized learning strategy, \emph{FedER}, that, exploiting experience replay and generative adversarial concepts, effectively integrates features from local nodes, providing models able to generalize across multiple datasets while maintaining privacy.
FedER is tested on two tasks --- tuberculosis and melanoma classification --- using multiple datasets in order to simulate realistic \emph{non-i.i.d.} medical data scenarios. Results show that our approach achieves performance comparable to standard (non-federated) learning and significantly outperforms state-of-the-art federated methods in their centralized (thus, more favourable) formulation. Code is available at \url{https://github.com/perceivelab/FedER}
\end{abstract}

\keywords{Decentralized Learning \and Federated Learning \and Privacy in Machine Learning \and Pattern recognition and classification}

\section{Introduction}
\label{sec:introduction}
Recent advances of deep learning in the medical imaging domain have shown that, while data-driven approaches represent a powerful and promising tool for supporting physicians' decisions, the availability of large-scale datasets plays a key role in the effectiveness and reliability of the resulting models~\cite{irvin2019chexpert,wang2017chestx,cohen2020covidProspective}. However, the curation of large medical imaging datasets is a complex task: data collection at single institutions is relatively slow and the integration of historical data may require significant efforts to deal with different data formats, storage modalities and acquisition devices; moreover, medical institutions are often reluctant to share their own data, due to privacy concerns. As a consequence, this affects the quality, reliability and generalizability of models trained on local datasets, which unavoidably suffer from bias and overfitting issues, reducing the ability to address future data distribution shifts~\cite{zech2018variable}.
In order to overcome the lack of large-scale datasets, methodological solutions can be adopted: in particular, federated learning~\cite{yang2019federated} encompasses a family of strategies for distributed training over multiple nodes, each with its own private dataset, which typically communicate with a central node by sending local model updates, used to train the main model. In this scenario, no data is explicitly shared between nodes, thus addressing the required privacy issues. 
However, this family of techniques generally performs well when dataset distributions are approximately \emph{i.i.d.} and local gradients/models contribute to learning shared features: unfortunately, in practice this hypothesis rarely holds, due to differences in the acquisition and in the clinical nature of data collected by multiple institutions. Moreover, the presence of a central node, besides representing a single point of failure, requires that all nodes trust it to correctly and fairly treat updates from all sources: indeed, privacy issues arise when transferring local updates to the ``semi-honest'' central node~\cite{evans2018pragmatic}, which might attempt to reconstruct original inputs from gradients or parameter variations~\cite{zhu2019deep,geiping2020inverting,zhao2020idlg}. 
To address the above limitations, we present \emph{FedER}, a federated learning approach that, leveraging experience replay from continual learning~\cite{ratcliff1990connectionist,robins1995catastrophic,rolnick2019experience,buzzega2020dark} and generative models~\cite{goodfellow2014generative,mirza2014conditional,karras2020analyzing},  proposes a principled way for training local models that approximately converge to the same decisions, without the need of a shared model architecture and of central coordination.
\emph{FedER} also enforces privacy preservation through the transmission of synthetic data generated in a way to obfuscate real data patterns.\\
Specifically, FedER's learning strategy envisages multiple nodes that initially train their local models and a GAN on their own datasets. The GAN will be used  in order to generate a privacy-preserving synthesized version of the dataset (buffer). Once local training is completed in a node, its model and the ``buffer'' of generated synthetic data are sent to a random node of the network. The receiving node then adapts the incoming model using its own data and the received buffer data, in order to limit model's forgetting. 
Data privacy is ensured through a privacy-preserving generative adversarial network (GAN) that employs a specific loss designed to maximize the distance from real data, while keeping a high level of realism and --- as importantly --- clinically-consistent features, in order to allow models to be trained effectively.

FedER is tested on two tasks, simulating a \emph{non-i.i.d.} medical scenario: 1) classification of tuberculosis from X-ray data, using Montgomery County and Shenzhen Hospital datasets~\cite{candemir2013lung,jaeger2014two,jaeger2013automatic}, and 2) melanoma classification using skin images of the ISIC 2019 dataset~\cite{combalia2019bcn20000,tschandl2018ham10000,codella2018skin}. The experimental setting is specifically designed to emulate a realistic medical \emph{non-i.i.d.} scenario, where each node in the federation uses its own dataset. This is in stark contrast with common procedures where non-\emph{i.i.d.} distributions are simulated by splitting a single source dataset. Results show how our approach is able to reach performance similar to using centralized training on all real data together in a single node, while outperforming current state-of-the-art methods, such as FedAvg~\cite{mcmahan2017communication}, FedProx~\cite{li2020federated} and FedBN~\cite{li2021fedbn}. Privacy-preserving capabilities are measured quantitatively by evaluating LPIPS distance~\cite{zhang2018unreasonable} between real images and samples generated, respectively, through latent space optimization on a standard GAN and by the proposed approach. Qualitatively, we also show several examples of generated images with corresponding closest match in the real dataset, demonstrating significant differences that prevent tracing back to the original real distribution.

In summary, the overall contributions of the proposed work are the following: 
\begin{itemize}
    \item We propose a decentralized federated learning strategy, based on continual learning principles, designed for medical imaging data, which outperforms centralized federated learning approaches and yields performance similar to standard (non-federated) training settings. Furthermore, experience replay allows local node models to converge to the same decisions, thus making the whole approach behave similarly to centralized aggregation models.
      \item We propose a GAN-based privacy-preserving mechanism that supports synthetic data sharing through a GAN-based technique designed to minimize patient information leak. This is different from most privacy-preserving techniques based on differential privacy, which degrades performance due to added noise.
    \item Most approaches for model aggregation in federated learning employ gradient/parameter averaging. These solutions completely neglect any similarity or dissimilarity between merged features, possibly resulting in interference that harm convergence. FedER, instead, takes feature semantics into account when merging models: if a node receives a model that extracts useful features for the local dataset, these can be readily employed and re-used, without the risk of randomly averaging them with other less important features.
    \item \emph{FedER}, thus, surpasses the common and straightforward weight/gradient averaging paradigm, replacing it with a principled way for knowledge transfer, which relaxes two of the constraints of the leading federated learning approaches: the presence of a central node and model homogeneity.

\end{itemize}

\section{Related Work}
\label{sec:related}
\begin{figure*}[h!]
\centering
\includegraphics[width=0.95\textwidth]{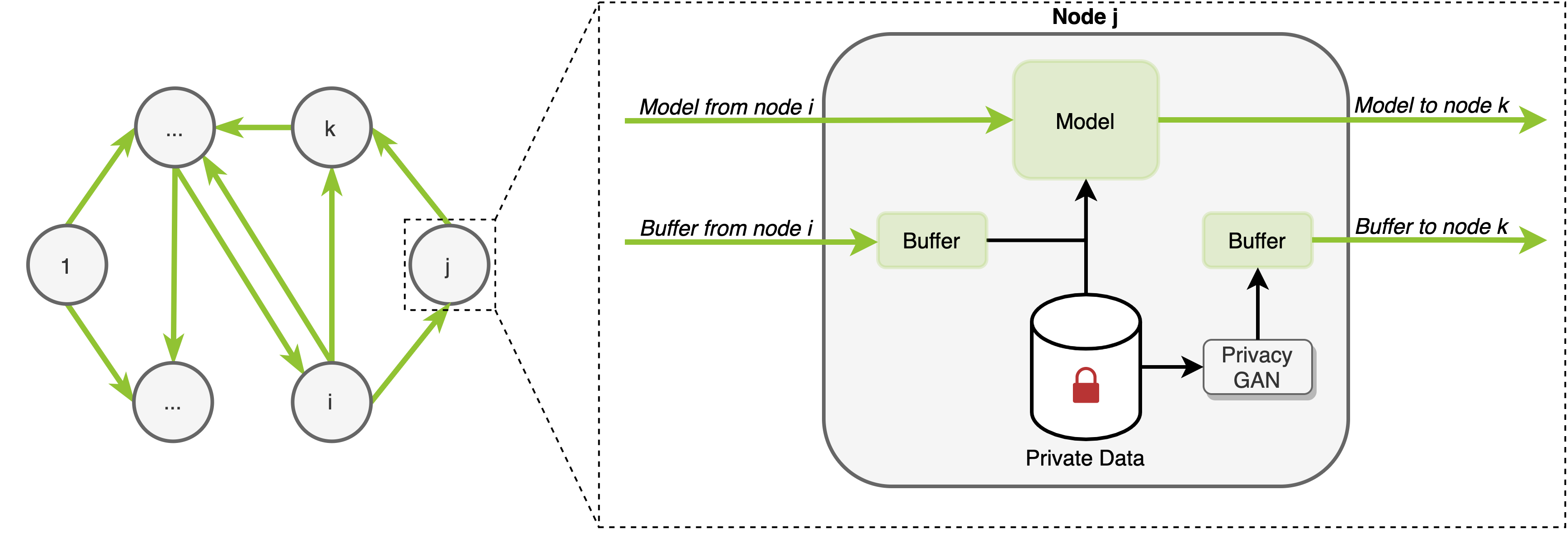}
\caption{\textbf{Overview of FedER  learning strategy.} Each node initially trains a \emph{privacy-preserving GAN}, that is used to sample synthetic data from the local distribution, without retaining features that may be used to identify patients. Then, each node iteratively receives the local model and a buffer of synthetic samples from a random node, and fine-tunes the received model on its own private data, using the buffer to prevent forgetting of previously-learned features.}
\label{fig:architecture}
\end{figure*}

Federated Learning (FL)~\cite{mcmahan2017communication} has recently emerged as a family of distributed learning strategies that allow nodes to keep training data private, while supporting the creation of a shared model. In a typical FL setting, a central server sends a model to a set of client nodes; each node fine-tunes the model on its own data, then sends local model updates back to the server; the server aggregates the updates by all nodes into the global model, which is sent back to nodes iteratively until convergence. Given the constraints existing in the medical domain, especially in terms of data sharing, it represents an appropriate test-bench for federated learning methods~\cite{li2019privacy,roy2019braintorrent,dayan2021federated,feki2021federated}. %
The most straightforward way to aggregate information from multiple nodes is through averaging local models of each client, as proposed in FedAvg~\cite{mcmahan2017communication} and FedProx~\cite{li2020federated}. However, statistical data heterogeneity is an issue as it may lead to catastrophic forgetting~\cite{kairouz2021advances,goodfellow2013}. FedCurv~\cite{shoham2019overcoming} addresses this limitation by adding a penalty term to the loss driving the local models to a shared optimum. FedMA~\cite{wang2020federated} builds a shared global model in a layer-wise manner by matching and averaging hidden elements with similar feature extraction signatures. 
Our method differs from existing feature integration approaches in that, instead of averaging model updates or gradients, which can be subject to input reconstruction attacks~\cite{geiping2020inverting,xie2018differentially,zhu2019deep}, each node attempts to learn features that perform well on its own dataset while retaining knowledge from other nodes, in a more principled way than parameter averaging.
The strategy of fitting the global model to local data is also sought by the recent federated \textit{personalized methods}. FedBN~\cite{li2021fedbn}, for instance, keeps batch normalization layers private, while other model parameters are aggregated by the central node.

However, the presence of a central node that aggregates local updates simplifies the communication protocol when the number of clients is very large (thousands or millions), but introduces several downsides: it represents a single point of failure; it can become a bottleneck when the number of clients increases~\cite{lian2017can}; in general, it may not always be available or desirable in collaborative learning scenarios~\cite{kairouz2021advances}. %
 In this paper, we deal with \emph{decentralized federated learning}, in which the central node is replaced by peer-to-peer communication between clients: there is no longer a global shared model as in standard FL, but the communication protocol is designed so that all local models approximately converge to the same solution. Decentralized learning is particularly suitable to application in the medical domain, where the number of nodes (i.e., institutions) is relatively low; however, research is still ongoing, and no effective solutions have been established. In~\cite{lalitha2019peer}, a Bayesian approach is proposed to learn a shared model over a graph of nodes, by aggregating information from local data with the model of each node's one-hop neighbors. A secure weight averaging algorithm is proposed in~\cite{wink2021approach}, where model parameters are not shared between nodes, but all converge to the same numerical values (with the disadvantages associated to parameter averaging with \emph{non-i.i.d.} data distributions). Other approaches implement different communication strategies based on parameter sharing (e.g., decentralized variants on FedAvg~\cite{sun2021decentralized,mcmahan2017communication}).  %
In general, many of the existing solutions do not target, nor are they tested on, the medical domains --- most employ toy datasets, such as MNIST and CIFAR10. Two works, similar in the decentralized learning spirit to ours, are proposed in ~\cite{roy2019braintorrent,9943283}, where use cases of decentralized and swarm learning for medical image segmentation are presented. However, like other approaches, they adopt simple parameter averaging to integrate features or predictions from multiple nodes.

\section{Method}
\label{sec:method}
\subsection{Overview}

An overview of FedER is shown in Fig.~\ref{fig:architecture}. In this scenario, a \emph{federation} consists of a set of $N$ peer nodes, each owning a private dataset.

Before the decentralized training algorithm is started, each node internally trains a \emph{privacy-preserving generative adversarial network}, which is used to generate synthetic samples from its private data distribution. The training objective of the GAN is designed to enforce the constraint that sampled data do not include privacy-sensitive information, while maintaining the clinical features required for successful training.

At each round of decentralized training, each node receives a model and a set of synthetic samples  --- ``buffer'' --- from a random node in the federation. The input model to the node is fine-tuned on both the private dataset and the buffer, in a way that is reminiscent of experience replay techniques in continual learning (e.g.,~\cite{buzzega2020dark}), in order to learn features that transfer between nodes and that can handle non-\emph{i.i.d.} distributions. At the end of each round (i.e., after performing several training iterations), the locally-trained model is sent to a randomly-chosen successor node together with a buffer of local synthetic samples, and the whole procedure is repeated.

In this work we specifically address the problem of federated learning for medical image classification; thus, the method is presented by considering this task, but the whole  strategy can be applied to any other task without losing generalization. 

\subsection{Privacy-preserving GAN}
\label{sec:ppgan}

In the proposed method, nodes exchange both models and data, implementing a knowledge transfer procedure based on experience replay (see Sect.~\ref{sec:exp_replay} below). Of course, sharing real samples would go against federated learning policies; hence, exchanged samples are generated so that they are representative of the local data, while taking precautions against privacy violations --- which may happen, for instance, if the generative model overfits the source dataset.

Formally, we assume that each node $n_i$, from a set of $N$ nodes, owns a private dataset $\dD_i = \left\{ \left(\xx_1, \yy_1\right), \left(\xx_2, \yy_2\right), \dots, \left(\xx_M, \yy_M\right)\right\}$, where each $\xx_j \in \xX$ represents a sample in the dataset, and each $\yy_j \in \yY$ represents the corresponding target\footnote{The proposed approach is task-agnostic, as long as it is possible to sample from the $\yY$ distribution. For simplicity, within the scope of this work, we will focus on classification tasks, and we will assume that targets are class labels.}. The local dataset can then be used to train a conditional GAN~\cite{mirza2014conditional}, consisting of a generator $G$, that synthesizes samples for a given label by modeling $P\left( \xx | \yy, \zz \right)$), where $\zz \in \zZ$ is a random vector sampled from the generation latent space, and a discriminator $D$, which outputs the probability of an input sample being real, modeling $P\left( \text{real} | \xx, \yy\right)$. The standard GAN formulation introduces a discrimination loss, which trains $D$ to distinguish between real and synthetic samples:
\begin{equation}
\lL_D = -\Ee_{\xx,\yy}\left[ \log\left( D\left( \xx, \yy \right)  \right)\right] - \Ee_{\zz,\yy} \left[ \log \left( 1 - D\left( G \left( \zz, \yy \right) , \yy \right) \right) \right],
\end{equation}
and a generation loss, which trains $G$ to synthesize samples that appear realistic to the discriminator:
\begin{equation}
\label{eq:g_loss}
\lL_G = -\Ee_{\zz,\yy} \left[ D\left( G\left(\zz, \yy\right), \yy \right) \right] .
\end{equation}

While it has been theoretically proven that, at convergence, the distribution learned by the generator matches and generalizes from the original data distribution~\cite{goodfellow2013}, unfortunately GAN architectures may be subject to training anomalies, including mode collapse and overfitting: as a consequence, the basic GAN formulation may lead to the generation of samples that are near duplicates of the original samples, which would be unacceptable in a federated learning scenario.

In order to mitigate this risk, we introduce a \emph{privacy-preserving loss}, enforcing the generation of samples that do not retain potentially sensitive information, but still include features that are clinically relevant to the target $\yy$ of the synthetic sample. In other words, if $\yy$ encodes generic features for the diagnosis of a certain disease, we want the generator to learn how to synthesize samples conditioned by $\yy$, that exhibit evidence of that disease but cannot be traced back to any of the dataset's samples of the same disease.

To do so, our privacy-preserving loss aims at penalizing the model proportionally to the similarity between pairs of real and synthetic samples. We measure ``similarity'' by means of the LPIPS metric~\cite{zhang2018unreasonable}, which has been shown to capture perceptual similarity by calibrating the distance between feature vectors extracted from a pre-trained VGG model~\cite{simonyan2015very}.

In practice, given a batch of real samples $\left\{ \xx_1^{(r)}, \xx_2^{(r)}, \dots, \xx_b^{(r)} \right\}$ and a batch of synthetic samples $\left\{ \xx_1^{(s)}, \xx_2^{(s)}, \dots, \xx_b^{(s)} \right\}$, the privacy-preserving loss term is computed as:
\begin{equation}
\label{eq:pp_loss}
\lL_\text{PP} = \frac{1}{b} \sum_{\xx^{(r)}}\sum_{\xx^{(s)}} d_L\left(\xx^{(r)}, \xx^{(s)}\right),
\end{equation}
where $d_L$ is the LPIPS distance. Note that, in this formulation, we ignore the $\yy$ targets associated to each $\xx$: we want to prevent the model from generating near-duplicates of real samples in general, regardless of class correspondence. Also, we intentionally employ a pairwise metric on samples, rather than an aggregated metric such as Fr\^echet Inception Distance~\cite{heusel2017gans}, since we want to prevent similarity between samples, not between distributions, which would conflict with the GAN objective.

The resulting new loss for the Generator is a combination of Eq.~\ref{eq:g_loss} and Eq.~\ref{eq:pp_loss}: 
\begin{equation}
\lL_{G\text{-PP}} = \lL_{G} - \alpha \lL_\text{PP}
\end{equation}
where $\lL_\text{PP}$ is sign reversed as we want to maximize Eq.~\ref{eq:pp_loss}, while $\alpha$ is a hyperparameter used to balance the two terms.

The combined effect of the three loss terms --- $\lL_D$, $\lL_G$, $\lL_\text{PP}$ --- pushes the generator to explore the sample space to match the dataset distribution, while ``avoiding'' latent space mappings that would project to actual real samples.

\subsection{Federated learning with experience replay}
\label{sec:exp_replay}

Current approaches for federated learning are mostly based on parameter averaging (e.g., FedAvg), which is, however, a straightforward way to combine knowledge from multiple sources: feature locations are not aligned over different models and may be disrupted by updates, before slowly converging to consensus: hypothetically, two models could learn the same set of features at different locations of the same layer, to only have them cancel each other when averaging. In a decentralized scenario, this issue is even exacerbated, due to the lack of an entity that enforces global agreement on node features. 

In our approach, we address this problem by taking inspiration from continual learning strategies~\cite{delange2021continual} that learn how to perform a task with a non-\emph{i.i.d} data stream  without forgetting  previously-learned knowledge: as a consequence, models are encouraged to reuse and adapt features so that they can equally serve the current and previous tasks. Analogously, in the federated learning setting, the objective is to train a global model trained on disjoint non-\emph{i.i.d.} data distributions coming from different nodes.

Given these premises, we define a federated learning strategy where a node receives another node's model and surrogate data (generated through our privacy-preserving GAN) --- the \emph{``previous task''} --- and fine-tunes that model on its own private date --- the \emph{``current task''} --- while using received synthetic data as a reference to what is necessary to retain/adapt from the knowledge learned by the previous 
node. The idea is to build for each node a model able to tackle its internal data while not forgetting about the data seen in previous nodes/iterations.

We first introduce the terminology used in the method's description. In our approach, we define a \emph{set of $N$ tasks} $\tT = (T_1, T_2, \dots, T_N)$, where $T_i$ is the task to be solved within node $n_i$.\\

\textbf{Definition 1}. Task $T_i$ aims at optimizing a model $M_i$, parameterized by $\bm{\theta}_i$, on dataset $\dD_i$ residing on node $n_i$ and that cannot be shared to other nodes.\\

\textbf{Definition 2}. A buffer $\bB_i$ is a set of synthetic images, drawn from a latent space learned through a generative model $\gG_i$ using data $\dD_i$ available on node $n_i$.\\

\textbf{Definition 3}. Training is organized in parallel \textit{rounds}. At the end of round $r$, each node $n_i$ produces a model $M_i^r$ trained on dataset $\dD_i$ and on a buffer $\bB_j$, received from another node $n_j$,
to optimize an objective $\lL$, i.e., to find $\argmin_{\bm{\theta}_i^r} = \EX_{(\xx,\yy)\sim \dD_i \cup \bB_j}[\lL(M_i^r(\xx,\bm{\theta}_i^r),\yy)]$. For each training round, all nodes in parallel share to/receive from other nodes, buffer of synthetic images and trained models. \\

In the following, we describe our method (whose graphical representation is given in Fig.~\ref{fig:architecture}) from the point of view of a single node $n_j$. At a given round $r$, training for node $n_j$ can be seen as learning a new task $T_j$, from dataset $\dD_j$, in a continual learning setting by finetuning the incoming model $\mM_i^{r-1}$ (with parameters $\bm{\theta}_i^{r-1}$) on $\dD_j$ and on the incoming buffer $\bB_i$ in order to learn $T_j$ while mitigating the forgetting of $T_i$. Thus, unlike other federated learning approaches, each node does not have its own local model: as the decentralized learning strategy proceeds, a node iteratively receives a model from another node and updates it with local information, while preserving previously-learned knowledge, before sending it to the next node. 
Formally, the loss function for model $\mM_j^r$ in node $n_j$ at round $r$ is given as:

\begin{equation}
\label{eq:lambda}
\begin{split}
    \mathcal{L}(\bm{\theta}_j^r) & = \lambda \EX_{(\xx,\yy)\sim \dD_j}[\mathcal{L}(M_j^r(\xx,\bm{\theta}_j^r),\yy)] + \\ 
    & + (1-\lambda) \EX_{(\xx',\yy')\sim \bB_i}[\mathcal{L}(M_j^r(\xx',\bm{\theta}_j^r),\yy')]
\end{split}
\end{equation}

where $\lambda$ controls the importance between real samples from the local dataset $D_i$ and replayed synthetic samples from node $n_i$.
Note that, for a given $n_j$, the predecessor node $n_{i}$ is not fixed: in a practical asynchronous implementation, a node may receive a model and buffer from any random node in the federation at any time, using queues to handle incoming data.

After optimizing the $\mathcal{L}(\bm{\theta}_j^r)$ objective through mini-batch gradient descent for a certain number of training iterations, the resulting model $M_j^r(\bm{\theta}_j^r)$, with updated parameters $\bm{\theta}_j^r$, is sent to a random node $n_k$ of the federation, along with a buffer $\bB_j$ of locally-generated synthetic samples. The number of training rounds/iterations and the size of the buffer is discussed in the next section.

Then, the general federated model $\mM$, after all training rounds, is given by the union of all the $N$ node models, i.e., $\mM = M_1 \cup M_2 \cup \cdots \cup M_N$. However, experimental results, reported below in Sect.~\ref{sec:performance}, demonstrate that all models converge to similar decisions, thus each node model can be considered as a general model for the entire network.

To ease the understanding of the whole training strategy we also report the algorithm pseudo-code in Alg.~\ref{alg:two}.

\begin{algorithm}
\DontPrintSemicolon 
\caption{FedER Learning Procedure}\label{alg:two}
\SetKw{KwNot}{Notations}
\SetKwComment{Comment}{/* }{ */}
\KwNot{The $N$ nodes are indexed by $n_i$;        
$E$ is the number of local epochs for each round.
$R$ the total round of communications between nodes.\\
Each node $n_i$ contains:\\
$\dD_i$ Private Dataset\\
$\gG_i$ Generator (privacy-preserving) trained on $D_i$\\
$\mM_i^r$ Model for node $n_i$ at round $r$\\
$\bB_i$ Synthetic data buffer sampled using $G_i$\\
}\;
 \tcp{Before Federated Training}
 \For{each node $n_i \in N$}{
 \textbf{Train} $\gG_i$ on $\dD_i$\\
 \textbf{Generate} Buffer $\bB_i$ using $\gG_i$\\
 \textbf{Train} $\mM_i^0$ on $\dD_i$\\
 }\;
 \tcp{Federated Training}
\For{each round $r = 1,2,...,R$}{
\For{each node $n_j \in N$ in parallel}{
    \textbf{Send} $\mM_j^{r-1}$ , $\bB_j$ to a node $n_k \in \{N \setminus n_j \} $\\
    \textbf{Receive} $\mM_i^{r-1}$, $\bB_i$ from a node $n_i \in \{N \setminus n_j \} $\\
    $\mM_j^{r} \gets \mM_i^{r-1}$\\
     \textbf{Train} $\mM_j^{r}$ on $\{\dD_j \cup  \bB_{i}\}$ for $E$ epochs\\
}
}
\end{algorithm}

\section{Experimental Results}
\label{sec:performance}

\begin{table*}[!h]

\centering
\caption{\textbf{Rounds and epochs in FedER.} Results (mean $\pm$ standard deviation) obtained with  5-fold cross-validation. Buffer size = 512.}
\label{tab:effectRoundEpoch}
\vspace{0.2cm}
\begin{tabular}{crccccc} 
\toprule
\multicolumn{1}{l}{}       &                      & \multicolumn{2}{c}{\textbf{Tuberculosis}}                                                               & \multicolumn{3}{c}{\textbf{Melanoma}}                                                                                      \\ 
\cmidrule(lr){3-4} \cmidrule(lr){5-7}
      &                      & \textbf{Shenzhen} & \textbf{Montgomery}  & \textbf{BCN} & \textbf{HAM}  & \textbf{MSK4} \\ 
 \cmidrule(lr){3-4} \cmidrule(lr){5-7}    
\multicolumn{1}{l}{Rounds} & Epochs & Accuracy & Accuracy & Accuracy & Accuracy & Accuracy \\
\midrule
\multirow{3}{*}{10}        %
                           & 1                    & 82.39 $\pm$ 6.91               & 56.13  $\pm$ 3.03 &     76.73 $\pm$ 2.07     &   82.24 $\pm$ 4.01                   &  67.93 $\pm$ 4.84  \\
                           
                           & 10                   & 82.86 $\pm$ 2.44               & 86.73  $\pm$ 4.22     & 83.83 $\pm$ 1.96     &    84.72 $\pm$ 2.29                   & 73.67  $\pm$ 2.59  \\
                           
                           & 100                  & 83.56 $\pm$ 1.72              & 90.79  $\pm$ 3.92     & 85.51 $\pm$ 1.85      & 88.65 $\pm$ 1.12                  & 71.81   $\pm$ 2.04 \\
\midrule
\multirow{3}{*}{100}       %
                           & 1                    & 83.31 $\pm$ 2.59              & 88.71  $\pm$ 3.82     & 78.94 $\pm$ 2.55      & 87.34 $\pm$ 1.62                   & 72.07  $\pm$ 3.45 \\
                           
                           & 10                   & 85.22 $\pm$ 2.42               & 89.72  $\pm$ 3.46     & 84.62 $\pm$ 1.40      & 85.05 $\pm$ 1.62                   & 73.72  $\pm$ 2.41 \\
                           
                           & 100                  & 87.10 $\pm$ 2.31               & 91.50  $\pm$ 2.60    & 86.06 $\pm$ 0.96      & 89.26 $\pm$ 1.11                   & 72.41  $\pm$ 1.53 \\ 
\bottomrule
\end{tabular}

\end{table*}

We test FedER on two applications simulating real case scenarios with multiple centers holding, and not sharing, their own data: 1) tuberculosis classification from X-ray images using two different datasets, and 2) skin lesion classification with three different datasets. In this section we present the employed benchmarks, the training procedure and report the obtained results to demonstrate the advantages of the proposed approach w.r.t. the state-of-the-art. 

\subsection{Datasets}
\label{sec:dataset}
\noindent \textbf{X-ray image datasets for tuberculosis classification}. We assume two separate nodes in the federation: one with the Montgomery County X-ray set and another one with the Shenzhen Hospital X-ray set~\cite{candemir2013lung, jaeger2014two, jaeger2013automatic}. The Montgomery Set consists of 138 frontal chest X-ray images (80 negatives and 58 positives), captured with a Eureka stationary machine (CR) at 4020$\times$4892 or 4892$\times$4020 pixel resolution. The Shenzhen dataset was collected using a Philips DR Digital Diagnostic system. It includes 662 frontal chest X-ray images (326 negatives and 336 positives), with a variable resolution of approximately 3000$\times$3000 pixels.

\noindent \textbf{Skin lesion classification}. We employ the ISIC 2019 challenge dataset, which contains 25,331 skin images belonging to nine different diagnostic categories. In this case, we assume a federation with three nodes as data provided belongs to three different sources: 1) the BCN20000~\cite{combalia2019bcn20000} dataset, consisting of 19,424 images of skin lesions captured from 2010 to 2016 in the Hospital Clínic in Barcelona; 2) the HAM10000 dataset~\cite{tschandl2018ham10000}, which contains 10,015 skin images collected over a period of 20 years from two different sites, the Department of Dermatology at the Medical University of Vienna, Austria, and the skin cancer practice of Cliff Rosendahl in Queensland, Australia; 3) the MSK4~\cite{codella2018skin} dataset, which is anonymous and includes 819 samples. Among all skin lesion classes, we only consider the melanoma class, posing the problem as a binary classification task.\\
In all tasks and datasets we adopt 80\% of the available data to train both the privacy-preserving GAN and the classification model, while the remaining 20\% of each dataset is used as test set. Test sets are also balanced w.r.t. the label to avoid performance biases due to class imbalance. 
For all tested federated methods (including state-of-the-art ones), model selection is carried out through with 5-fold cross-validation on the training set, as a grid search on number of training rounds, number of rounds per epoch and learning rate. For FedProx~\cite{li2020federated}, we also include the $\mu$ hyperparameter.

\subsection{Training procedure and metrics}

\subsubsection{Federated training}

In all settings, we employ ResNet-18 as classification model, trained by minimizing the cross-entropy loss with mini-batch gradient descent using the Adam optimizer. Mini-batch size is set to 32 and 8 for the Shenzhen and Montgomery datasets, respectively, and to 64 for skin lesion datasets.
The learning rate was found, through cross-validation, to be $10^{-4}$. Data augmentation is carried out with random horizontal flip; for skin images we additionally apply random 90-degree rotations. All images are resized to 256$\times$256.
The ratio between real and synthetic samples controlled by $\lambda$ in Eq.~\ref{eq:lambda} is set to $0.5$ for all experiments, i.e., each mini-batch is composed by the same quantity of real and synthetic images. This also ensures that our method performs the same number of optimization steps as other approaches that do not use any synthetic data.

The node federation is trained for $R$ rounds. %
In our implementation, at each round nodes are randomly ordered to establish each node's predecessor and successor: given our focus on medical applications, we can assume that the number of nodes is low enough that synchronization is not an issue. However, asynchronicity can be achieved by assuming that nodes can store incoming data in a queue: if the distribution of successor nodes is uniform and computation times are similar for all nodes, this is on average equivalent to the synchronous case.
The number of rounds $R$ and epochs $E$ for FedER on the tuberculosis and melanoma classification tasks are set both to 100, according the 5-fold cross-validation results shown in Table~\ref{tab:effectRoundEpoch}. Buffer size is set for all experiments to 512.

\begin{table*}[h!]
\centering
\caption{\textbf{Comparison between FedER and centralized baselines}. Results for FedER are obtained with a buffer size of 512, 100 rounds and 100 epochs per round.}
\label{tab:upper_bound}
\vspace{0.2cm}

\resizebox{\linewidth}{!}{%
\begin{tabular}{lcccccccc} 
\toprule
\multirow{2}{*}{\textbf{Methods}}        & \multicolumn{3}{c}{\textbf{Tuberculosis}}      & \multicolumn{4}{c}{\textbf{Melanoma}}  \\ 
\cmidrule(lr){2-4} \cmidrule(lr){5-8}
                                & \textbf{Shenzhen} & \textbf{Montgomery} & \textbf{Mean}  & \textbf{BCN}  & \textbf{HAM}   & \textbf{MSK4}  & \textbf{Mean}\\ 
\toprule
\red{Standalone}  & \red{82.31}  & \red{90.00}  & \red{86.16}          & \red{82.90} & \red{82.55} & \red{69.75} & \red{78.40} \\
\midrule
Centralized training         & \red{82.77}   & \red{77.67}       & \red{80.22}     &  \red{78.80} & \red{82.90} & \red{71.23} & \red{77.64} \\
Centralized training with synthetic data only     & \red{76.92}   & \red{79.33}   & \red{78.13}      & \red{60.71} & \red{61.09} & \red{61.23} & \red{61.01}  \\
\red{Centralized training with synthetic data and real data}  & \red{85.38}   & \red{86.67}   & \red{86.03}      & \red{81.53} & \red{80.44} & \red{73.46} & \red{78.48}  \\
\midrule
\emph{FedER} (ours)         & \red{80.15}   & \red{86.67} & \red{83.41}    & \red{82.11}      & \red{84.58}     & \red{68.40} & \red{78.36} \\
\bottomrule
\end{tabular}}
\end{table*}

\subsubsection{GAN training}
We recall that GAN training is carried out \textbf{before} federated learning using training data only, while leaving out test samples, as mentioned in Sect.~\ref{sec:dataset}.
Our privacy-preserving GAN employs StyleGAN2-ADA~\cite{NEURIPS2020_8d30aa96} as a backbone, because of its suitability in low-data regimes and its generation capabilities. Training is carried out in two steps: 1) the GAN is initially trained without any privacy-preserving loss to support learning of high-quality visual features; 2) afterwards, we enable privacy-preserving loss and fine-tune the model in order to limit the embedding of patient-specific patterns in the GAN latent space.
For classification purposes, GANs are trained in a label-conditioned fashion with a mini-batch size of 32 and learning rate of 0.0025 for both the generator and the discriminator. Early-stopping criteria are based on the Fr\^echet Inception Distance (FID)~\cite{heusel2017gans} between real and synthetic distributions: in the first training step, we stop training if FID does not improve for 10,000 iterations; in the second training step, we employ a criterion which stops training if FID increases by a factor of 2.5 w.r.t. the value obtained in the first step. 
As for the $\alpha$ parameter in Eq.~\ref{eq:pp_loss}, we tested multiple values of $\alpha$ (0, 0.5, 1, 1.5, 2 and 3) and found that the value of 1 yields the best compromise between image generation quality and pairwise LPIPS distance~\cite{zhang2018unreasonable} over all tested datasets. %
In order to quantitatively evaluate privacy preservation, we also compute the average LPIPS distance between each real image and its closest synthetic sample by means of latent space projection (described in Sect.~\ref{sec:proj}): the higher value of LPIPS, the lower the possibility to reconstruct real images from the generator.

\begin{table}[h!]
\centering
\caption{\textbf{Accuracy convergence among distributed node models}.Each local model is evaluated on all test sets of the federation in order to measure convergence and generalization (lower standard deviation corresponds to higher convergence).}
\label{tab:convergence}

\begin{tabular}{clcc}
\toprule
                                                           &        \textbf{Dataset}             & \textbf{FedER} & \textbf{Standalone}  \\ 
    \toprule
\multicolumn{1}{c}{\multirow{2}{*}{\textbf{Tuberculosis}}} & \textbf{Shenzhen}      & $80.54 \pm 1.20$                        &     $66.15 \pm 22.84$         \\
\multicolumn{1}{c}{}                                       & \textbf{Montgomery} & $85.67 \pm 2.36$                        &    $70.00 \pm 28.28$   \\ 
\midrule
\multirow{3}{*}{\textbf{Melanoma}}                         & \textbf{BCN}        & $82.87 \pm 1.22$                     &     $65.06 \pm 19.68$    \\
                                                           & \textbf{HAM}        & $84.45 \pm 0.75$                      &  $59.94 \pm 20.47$  \\
                                                            & \textbf{MSK4}       & $67.78 \pm 1.28$                   &   $65.43 \pm 5.05$  \\ 
\bottomrule
\end{tabular}
\end{table}

\subsection{Federated learning performance}
We first evaluate the performance (in terms of classification accuracy) of FedER in the non-\emph{i.i.d.} setting, and compare it to several centralized baselines, namely:
\begin{itemize}
    \item \textbf{Centralized training}: all datasets are merged in a single node where all training happens. In this setting, no federated learning constraints are applied. 
    \item \textbf{Centralized training with synthetic data only}. In this setting, each node trains a privacy-preserving GAN model and shares a synthetic version of its own data with the central node, where global training is performed. In this case, we aim to assess how much information is retained by synthetic data to support classification. 
    \item \textbf{Centralized training with synthetic and real data.} This setting is a combination of the previous two: real and synthetic samples are centrally merged and used for training a global classifier. This scenario measures the contribution of synthetic data as a data augmentation approach.   
\end{itemize}

We also compare FedER against standard training of the local node models, referred to as ``Standalone'' . Classification accuracy is computed using local node models on their own data. The results, reported in Table~\ref{tab:upper_bound}, show that standalone training appears to be the most favourable scenario. Centralized strategies perform generally worse than standalone training, because of the non-\emph{i.i.d.} nature of the data. However, when the centralized approach is trained with original data augmented with synthetic samples, its classification accuracy is on par with the standalone training, possibly due to the learned generative latent spaces that likely tend to smooth different modes of non-\emph{i.i.d.} data. FedER, instead, outperforms its centralized counterpart and yields slightly worse performance (1.5 percent points less) than standalone training. Although this may appear, at a first glance, as a shortcoming of FedER, we recall that in a federated learning scenario, we aim at building a model that, leveraging multiple data distributions present in the federation, may generalize better, thus addressing possible future data drifts. In order to assess the capabilities of the trained models to achieve such a generalization, we measure the decision convergence by evaluating how a local node model performs on other node datasets. Results are in Table~\ref{tab:convergence} and show a good average accuracy, with a low standard deviation, by FedER, indicating that each node model performs equally well on its own dataset and on the others (i.e., all node models converge to similar decisions). Conversely, standalone training yields significantly lower accuracy and higher standard deviation than ours, demonstrating to be an unsuitable strategy for the sought generalization properties.

\begin{table*}
\centering
\caption{\textbf{Comparison with state-of-the-art methods}. In bold, best accuracy values.}
\vspace{0.2cm}
\label{tab:sota_noniid}

\begin{tabular}{rccccccc} 
\toprule
        & \multicolumn{3}{c}{\textbf{Tuberculosis}} & \multicolumn{4}{c}{\textbf{Melanoma}}   \\ 
\cmidrule(lr){2-4} \cmidrule(lr){5-8}
        & \textbf{Shenzhen} & \textbf{Montgomery} & \textbf{Mean}      & \textbf{BCN}   & \textbf{HAM}   & \textbf{MSK4}  & \textbf{Mean}   \\ 
\midrule
FedAvg~\cite{mcmahan2017communication}   & 72.31   & 83.33      & 77.82     & 77.55      &  75.15     &     67.28  &   73.33     \\
FedProx~\cite{li2020federated} & 78.46   & 76.67      & 77.56     &  78.80     &  81.87     & 64.81      &  75.16      \\
FedBN~\cite{li2021fedbn}   & 63.08   & 70.00      & 66.54     & \textbf{82.19} & 81.12 & 59.26 & 74.19  \\

\emph{FedER} (ours)    & \textbf{80.15}   & \textbf{86.67}      & \textbf{83.41}     &  82.11     &  \textbf{84.58}     &  \textbf{68.40}     &  \textbf{78.36}      \\
\bottomrule

\end{tabular}
\end{table*}

\begin{figure}[!h]
    \centering
    \includegraphics[width=\textwidth]{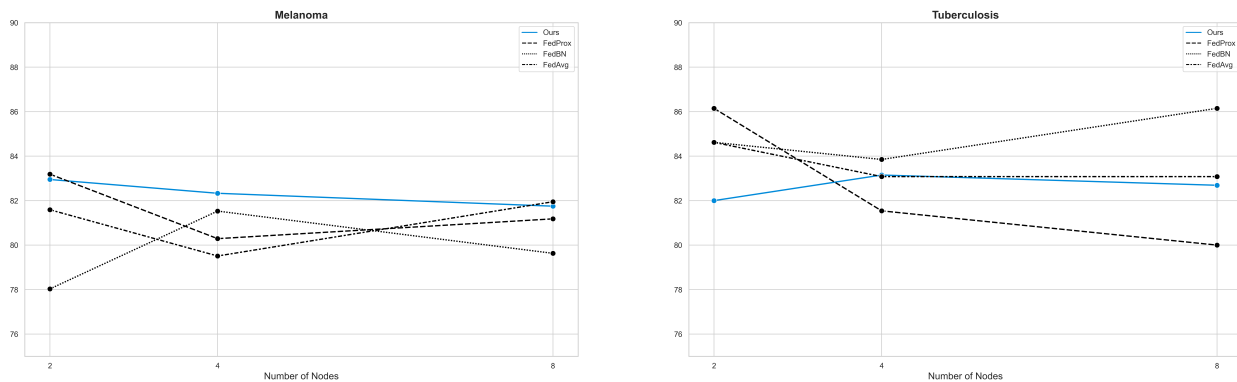}
    \caption{\textbf{Scalability performance in the \emph{i.i.d.} setting} w.r.t. number of nodes for the proposed approach and state-of-the-art methods.}
    \label{fig:iid}
\end{figure}

We then compare our approach to state-of-the-art federated learning approaches, namely: a) centralized federated methods, FedAvg~\cite{mcmahan2017communication} and FedProx~\cite{li2020federated},  which have shown to perform generally better than decentralized methods~\cite{sun2021decentralized,lalitha2019peer}, and b) a personalized method, FedBN~\cite{li2021fedbn}. 
As already mentioned, to avoid biased assessment, we use the official code repository\footnote{\url{https://github.com/med-air/FedBN}} of FedBN~\cite{li2021fedbn} and  hyper-parameter selection on the tested datasets was carried out through grid search on training rounds/epochs, learning rate and $\mu$ for FedProx~\cite{li2020federated} using 5-fold cross validation as for our approach.
Results, for the tuberculosis and the melanoma tasks, are reported in Table~\ref{tab:sota_noniid} and show that FedER outperforms all methods under comparison. Interestingly, FedER learning strategy does better than: a) \textit{centralized methods}, FedAvg~\cite{mcmahan2017communication} and FedProx~\cite{li2020federated}, suggesting that experience replay is a more effective feature aggregation approach than naive parameter averaging; b) personalized methods, such as FedBN~\cite{li2021fedbn}, which affects a limited aspect of feature representation (i.e., input layer distributions), while our approach adapts the entire model to local and remote tasks.\\
These above results suggest that experience replay plays a key role in federated models as a principled way to integrate features coming from different data distributions. To further assess its contribution, we evaluate FedER performance when using buffer at different sizes. Results on the tuberculosis task, measured as mean and standard deviation of the local node models over a given dataset, are shown in Table~\ref{tab:buffer} and indicate a clear contribution of the buffer in terms of overall performance and models' agreement. Indeed, with no buffer we obtain the lowest average performance and the highest standard deviation. As the buffer is enabled, we can observe a performance gain (mainly for the Shenzhen dataset) and a significant drop in standard deviation. Performance improves as buffer size increases, although gain becomes negligible above 512. Since higher buffer sizes result in more data to be shared among nodes, we use a buffer size of 512, as the best trade-off between accuracy and communication costs.

\begin{table}
\centering
\caption{\textbf{FedER classification accuracy w.r.t. buffer size}. Each local model is evaluated on all test sets of the federation in order to measure convergence and generalization (lower standard deviation corresponds to higher convergence).}
\label{tab:buffer}

\begin{tabular}{ccccc} 
\toprule
       &       \multicolumn{2}{c}{\textbf{Node Convergence}}   \\
            \cmidrule(lr){2-3} 
\textbf{Buffer}                            &       \textbf{Shenzhen}      & \textbf{Montgomery}   \\ 
\midrule
0    & ~$70.62 \pm11.97 $ & ~$80.33 \pm 10.84$ \\
256  & $80.46 \pm 2.96$   & $81.67 \pm 4.24$ \\
512 & $80.54 \pm 1.20$ & $85.67 \pm 2.36$\\
1024 & $82.23 \pm 1.31$ & $86.00 \pm 3.01$\\
2048 & $82.08 \pm 1.39$ & $88.67 \pm 2.97$\\
\bottomrule
\end{tabular}

\end{table}

We finally evaluate the capability of FedER to scale with the size of the federated network. Accordingly, we quantify this property using  an \emph{i.i.d.} setting on both tuberculosis (Shenzhen dataset) and skin lesion classification (BCN dataset) tasks, by equally splitting the available data on multiple nodes. Fig~\ref{fig:iid} shows how the proposed approach is able to keep classification accuracy high and performs on par with state-of-the-art approaches (namely, FedAvg, FedProx and FedBN). %

\subsection{Privacy-preserving performance}
\label{sec:proj}

In this section we quantify how much information of real samples is retained by our privacy-preserving method, and in particular in the mapping between latent space and synthetic images. To do so, we employ the projection method proposed in~\cite{karras2020analyzing}: given a real image $\xx$, we find an intermediate latent point $\ww$ such that the generated image $G(\ww)$ is most similar to $\xx$, by optimizing $\ww$ to minimize the LPIPS distance~\cite{zhang2018unreasonable} between $\xx$ and $G(\ww)$.

\begin{figure}[h!]
\centering
\includegraphics[width=0.44\textwidth]{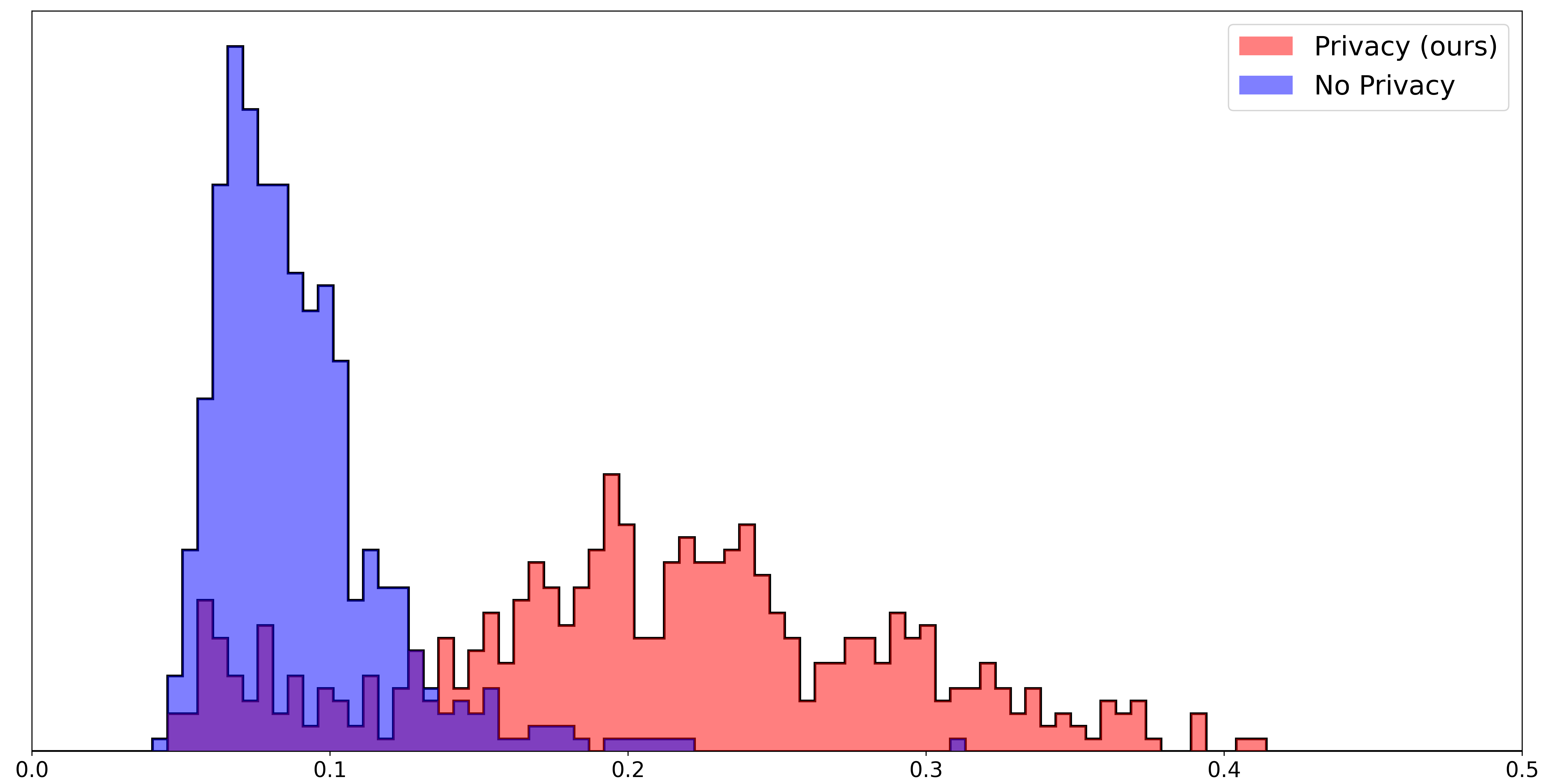}
\caption{\textbf{Quantitative analysis of privacy-preserving generation}. In blue, LPIPS distance histogram between real images and the corresponding images obtained through latent space projection using a GAN trained without the proposed privacy-preserving loss. In red, LPIPS distance histogram between real images and the closest images generated with the proposed approach.}
\label{fig:hist_privacy}
\end{figure}

\begin{figure*}[!h]
\centering
\includegraphics[width=0.95\textwidth]{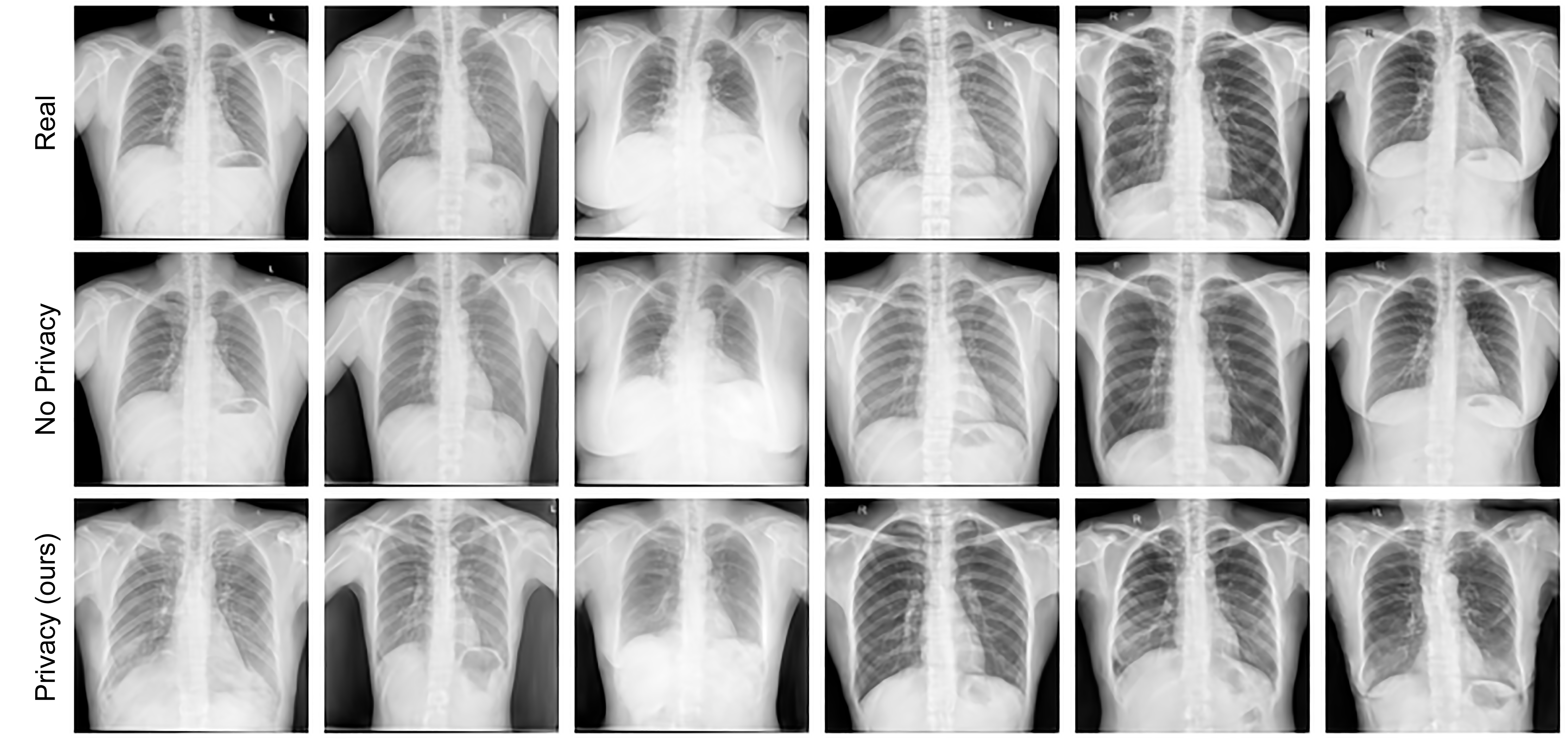}
\caption{\textbf{Qualitative samples of our privacy-preserving generation.} Top row: real images from the Shenzhen dataset. Middle row: projection with a standard GAN. Bottom row: projection with our privacy-preserving GAN.}
\label{fig:samples_privacy}
\end{figure*}

\begin{figure*}[h!]
\centering
\includegraphics[width=\textwidth]{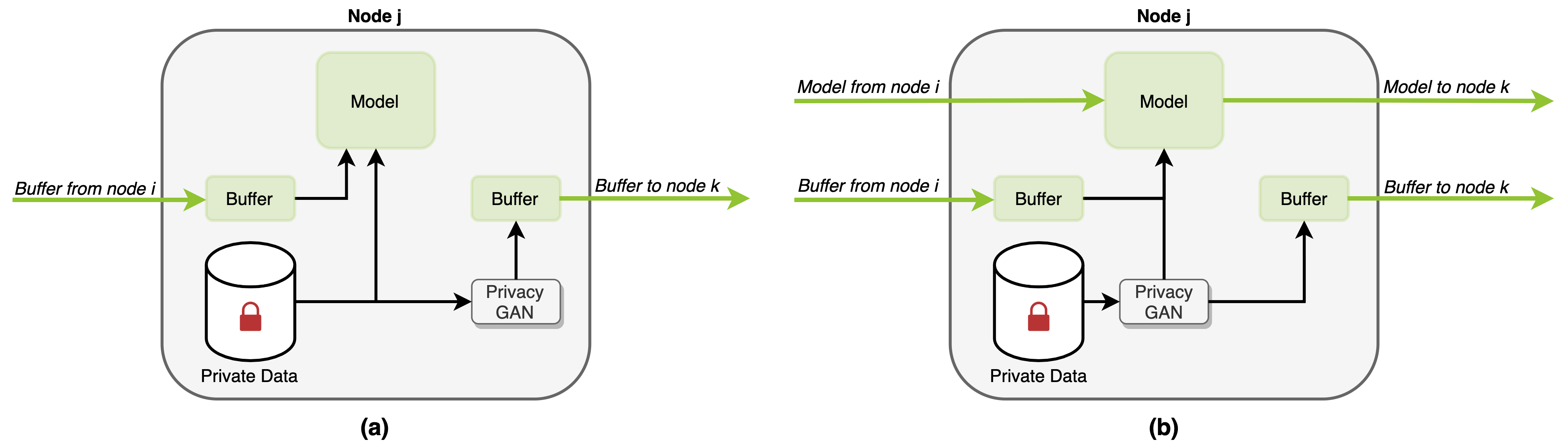}
\caption{\textbf{Privacy-enhanced alternative architectures.} \textit{(a)} \emph{FedER-A} configuration (``Buffer-only sharing''): a local node model is trained on real data, but only a buffer of synthetic samples is shared with other nodes. \textit{(b)} \emph{FedER-B} (``Synthetic-only training''): Even within the dataset owner node, models are trained on synthetic data only.}
\label{fig:privacy_enhanced_configs}
\end{figure*}

In practice, for each image of the dataset used for GAN training, we perform backprojection to find its most similar synthetic sample, and measure the LPIPS distance between the original and projected images. Fig.~\ref{fig:hist_privacy} shows the histograms of the resulting distances on the Shenzhen dataset, using GAN models trained with and without the proposed privacy-preserving loss (both models start from the same $\ww$, for fairness). The histograms show that standard GAN training, with no privacy-preserving loss, tends to yield distances closer to 0, demonstrating that real images are indeed included into the generator latent space; while our model significantly mitigates this issue, by synthesizing samples that are substantially different than the original ones. In order to qualitatively substantiate these findings, Fig.~\ref{fig:samples_privacy} compares original samples from the Shenzhen dataset with the corresponding projections, generated with and without our privacy-preserving loss\footnote{We show only X-Ray synthesized samples, as the effect is of our privacy-preserving strategy, is more appreciable than in skin lesion data.}. It is easy to notice that generated samples with a traditional GAN highly resemble real data, making it impossible to share such samples, albeit synthetic, in a privacy-safe manner, as they clearly contain patient information. 
Instead, comparing real images with the projections obtained from privacy-preserving GAN confirms the inability of the generator to find latent representations that recover real images used during training.

Given the high realism of generated samples, we run additional tests by proposing two FedER variants aiming to increase the level of privacy preservation: a) \emph{FedER-A}: models are not shared among nodes --- only synthetic buffers are sent and received; b) \emph{FedER-B}: models are trained only using synthetic data, even on local nodes. Fig.~\ref{fig:privacy_enhanced_configs} shows the internal architecture of each node in the two variants. Results obtained with these alternative privacy-enhanced configurations are provided in Table~\ref{tab:privacy_alternatives}. It can be noted that FedER-A (i.e., ``buffer-only sharing'') configuration achieves comparable performance to our standard FedER (82.76 vs 83.41), but, remarkably, it outperforms all existing federated learning methods on the same datasets (compare  Table~\ref{tab:sota_noniid} with the node performance block in Table~\ref{tab:privacy_alternatives}). The FedER-B (i.e., ``synthetic-only training'') configuration, instead, performs slightly worse than the other two configurations, but is still on par with existing federated methods. 

\begin{table}[h!]
\centering
\caption{\textbf{Classification accuracy of the proposed privacy-enhanced strategies in the \emph{non-i.i.d.} setting}.  FedER-A: only buffers are shared (Fig.~\ref{fig:privacy_enhanced_configs}-a). FedER-B: models are trained on synthetic data only (Fig.~\ref{fig:privacy_enhanced_configs}-b). Node performance measures how each node model performs on its own private dataset, while node convergence assesses how a node model performs on other federation nodes.}
\label{tab:privacy_alternatives}

\begin{tabular}{lccccc}
\toprule
\multirow{2}{*}{\textbf{Config}}&\multicolumn{3}{c}{\textbf{Node Performance}} & \multicolumn{2}{c}{\textbf{Node Convergence}}\\
\cmidrule(lr){2-4} \cmidrule(lr){5-6}
 & \textbf{Shenzhen} & \textbf{Montgomery}   & \textbf{Mean} & \textbf{Shenzhen} & \textbf{Montgomery} \\ \midrule
FedER                         & \red{80.15}  & \red{86.67}   & \red{83.41} & \red{80.54 $\pm 1.20$} & \red{85.67 $\pm 2.36$}\\
FedER-A              & \red{83.54}  & \red{82.00}   & \red{82.76} & \red{78.84 $\pm 6.64$} & \red{81.00 $ \pm 3.30$}\\
FedER-B      & \red{74.15}  &  \red{81.33}  & \red{77.74}& \red{73.61 $\pm 4.68$} & \red{80.40 $\pm 3.60$}\\ 
\bottomrule
\end{tabular}
\end{table}

\subsection{Communication and computational performance}

We conclude the experimental analysis by measuring \emph{communication} and \emph{computational} costs.

As for \emph{communication costs}, compared to state-of-the-art approaches, FedER requires additional transmission of synthetic images between nodes at each round. Tab.~\ref{tab:comm_costs_v2} reports per-node communication costs for state-of-the-art models (the table reports FedAvg, but the same values apply for FedProx and FedBN) and for FedER, in its full formulation and in the FedER-A variant, where only buffers of synthetic data are shared. The main cost for state-of-the-art models lies in the transfer of the model, and depends on the specific architecture (we included ResNet-18 and ResNet-152 as representative examples of different model scales). Values for our approach are reported for buffers of size 512 containing 256$\times$256 images, and depend on the color space. For our full FedER model, the increment in communication costs is significant but not excessive. However, if we take into consideration the variant where only synthetic data are exchanged (i.e., FedER-A), which still performs better than state-of-the-art methods (Tab.~\ref{tab:sota_noniid} and Tab.~\ref{tab:privacy_alternatives}), communication overhead becomes significantly less than model-sharing approaches.

As for \emph{computational costs} of federated training, FedER incurs the same overhead for parameter optimization and aggregation as state-of-the-art methods.
Additionally, before federated training starts, FedER requires that each node trains a local privacy-preserving GAN off-line; this, however, does not affect online federated learning costs, as it is carried out only once at the very beginning of the whole procedure.

Furthermore, we argue that, in the medical domain, the number of institutions in a federation is relatively low and it is reasonable to assume that nodes can benefit from a powerful communication network and computing infrastructure: thus, the overhead introduced by FedER is tolerable, in light of the methodological advantages and the obtained performance and generalization capabilities showed by the resulting models.

\begin{table}[h!]
\centering
\caption{\textbf{Communication results comparison}}
\label{tab:comm_costs_v2}

\begin{tabular}{lcccc} 
\toprule
        & \multicolumn{2}{c}{\textbf{Tuberculosis}} & \multicolumn{2}{c}{\textbf{Melanoma}}   \\ 
\cmidrule(lr){2-3} \cmidrule(lr){4-5}
        & \textbf{ResNet-18} & \textbf{ResNet-152}  & \textbf{ResNet-18} & \textbf{ResNet-152}  \\ 
\midrule
FedAvg   & \multirow{3}{*}{45 MB}   & \multirow{3}{*}{230 MB}      & \multirow{3}{*}{45 MB}     & \multirow{3}{*}{230 MB} \\
FedProx & & & &\\
FedBN & & & &\\
\midrule
FedER  & 65 MB  & 250 MB   & 105 MB    &  290 MB  \\
FedER-A   & 20 MB   & 20 MB     & 60 MB     &  60 MB     \\
\bottomrule
\end{tabular}
\end{table}

\section{Conclusion}
\label{sec:conclusion}
In this paper, we propose FedER, a decentralized federated learning framework that replaces traditional parameters averaging with a more principled feature integration approach based on the combination of experience replay and privacy-preserving generative models. In FedER, nodes communicate with each other by sharing local models and buffers of synthetic samples; local model updates are carried out in a way that encourages the reuse and adaptation of features learned by other nodes, thus avoiding potentially disruptive effects due to blind feature averaging. Experimental results show that our method outperforms significantly state-of-the-art centralized approaches in a non-\emph{i.i.d.} scenario, which is a typical setting in the medical domain. Additionally, quantitative and qualitative analysis shows that our privacy-preserving generation approach is able to synthesize samples that are significantly different from real data, while correctly supporting the learning of discriminative features. In the future, we aim at investigating some unexplored properties of our method: for instance, unlike all other existing methods based on parameter averaging is required, our approach does not strictly require that all nodes share the same model architecture. Model heterogeneity could therefore be employed to create a shared ensemble and combine different feature learning capabilities.

\end{document}